# Synchronized Stepwise Control of Firing and Learning Thresholds in a Spiking Randomly Connected Neural Network toward Hardware Implementation

**Kumiko Nomura*†, Yoshifumi Nishi†**

Frontier Research Laboratory, Corporate Research & Development Center, Toshiba Corporation, Kawasaki, Japan

† These authors contributed equally to this work and share first authorship

**\* Correspondence:**
Kumiko Nomura,
Frontier Research Laboratory, Corporate Research & Development Center, Toshiba Corporation, Komukai-Toshiba-cho 1, Saiwai-ku, Kawasaki 212-8582, Japan

kumiko.nomura@toshiba.co.jp

**Keywords: spiking neural network, intrinsic plasticity, synaptic plasticity, randomly connected neural network, neuromorphic chip**

## Abstract

We propose hardware-oriented models of intrinsic plasticity (IP) and synaptic plasticity (SP) for spiking randomly connected recursive neural network (RNN). Although the potential of RNNs for temporal data processing has been demonstrated, randomness of the network architecture often causes performance degradation. Self-organization mechanism using IP and SP can mitigate the degradation, therefore, we compile these functions in a spiking neuronal model. To implement the function of IP, a variable firing threshold is introduced to each excitatory neuron in the RNN that changes stepwise in accordance with its activity. We also define other thresholds for SP that synchronize with the firing threshold, which determine the direction of stepwise synaptic update that is executed on receiving a pre-synaptic spike. We demonstrate the effectiveness of our model through simulations of temporal data learning and anomaly detection with a spiking RNN using publicly available electrocardiograms. Considering hardware implementation, we employ discretized thresholds and synaptic weights and show that these parameters can be reduced to binary if the RNN architecture is appropriately designed. This contributes to minimization of the circuit of the neuronal system having IP and SP.

## 1    Introduction

Randomly connected recursive neural networks (RNNs), which have been studied as a simplified theoretical model of the nervous system of biological brains (Sompolinsky et al., 1988; Kadmon and Lazar et al., 2009; Somopolinsky, 2014; Bourdoukan et al., 2015; Tetzlaff et al., 2015; Thalmeier et al., 2016; Landau and Somopolinsky, 2018; Frenkel et al., 2022), are attracting much attention as a promising artificial intelligence (AI) technique that can perform prediction and anomaly detection of time series data in real time without executing sophisticated AI algorithms (Jaeger, 2001; Maass et al., 2002; Sussillo and Abbott, 2009; Nicola and Clopath, 2017; Das et al., 2018; Bauer et al., 2019). In particular, hardware implementation of RNNs is expected to reduce the power consumption of time series data processing, enabling intelligent operations of edge systems in our society. While the potential of RNNs has been well demonstrated in previous works (Jaeger, 2001; Maass et al., 2002; Sussillo and Abbott, 2009; Nicola and Clopath, 2017; Das et al., 2018; Bauer et al., 2019; Covi et al., 2021), inherent randomness sometimes causes uncontrollable data inference failures, leading to low reliability of the technique. Self-organization mechanism improves the reliability, which can be realized by including intrinsic plasticity (IP) and synaptic plasticity (SP) in the neuronal operation model (Lazar et al., 2009). IP is a homeostatic mechanism of biological neurons that controls neuron firing frequencies within a certain range. It has been shown to be indispensable for unsupervised learning in neuromorphic systems (Desai et al., 1999; Steil, 2007; Bartolozzi et al., 2008; Lazar et al., 2009; Diehl and Cook, 2015; Qiao et al., 2017; Davies et al., 2018; Payvand et al., 2022). SP is a mechanism where a synapse changes its own weight in accordance with incoming signals and the post-synaptic neuron's activity, known as the fundamental principle of learning in biological brains (Legenstein et al., 2005; Pfister et al., 2006; Ponulak and Kasinski, 2010; Kuzum et al., 2012; Ning et al., 2015; Prezioso et al., 2015; Ambrogio et al., 2016; Covi et al., 2016; Kreiser et al., 2017; Srinivasan et al., 2017; Ambrogio et al, 2018; Faria et al., 2018; Li et al., 2018; Amirshahi et al., 2019; Cai et al., 2020; Yongqiang et al., 2020; Dalgaty et al., 2021; Frenkel et al., 2023).

Since computing resources may be limited at the edge, we focus on analog spiking neural network (SNN) hardware having ultimately high-power efficiency for edge AI devices (Qiao et al, 2015; Davies et al., 2018; Payvand et al., 2022). The most general neuron model for SNNs is the leaky integrate-and-fire (LIF) model (Holt and Koch, 1997). For a LIF neuron, IP function may be added by adjusting its time constant of the membrane potential $V_{mem}$ according to its own firing rate $F_{fire}$. If we are to design LIF neurons with analog circuitry, tunable capacitor and resistor are required to control the time constant. The former is difficult because no practical device element having variable capacitance has been invented. For the latter, Payvand et al. (Payvand et al., 2022) proposed an IP circuit using memristors, namely, variable resistors. However, this circuit requires an auxiliary unit for memristor control, whose details are not yet discussed. Considering large device-to-device variability of memristors, each unit must be tuned according to the respective memristor's characteristics, which would result in a complicated circuit system with large overhead (Payvand et al., 2020; Demirag et al., 2021; Moro et al., 2022; Payvand et al., 2023).

Alternative method for controlling $F_{fire}$ is to adjust the firing threshold $V_{thr}$ itself (Diehl and Cook. 2015; Zhang and Li, 2019; Zhang et al., 2021). For a LIF neuron designed with analog circuitry, $V_{thr}$ is given as a reference voltage applied to a comparator connected to the neuron's membrane capacitor (Chicca et al., 2014; Qiao et al, 2015; Chicca et al., 2020; Payvand et al., 2022), hence IP can be implemented by adding a circuit that can change the reference voltage in accordance with $F_{fire}$. It would be straightforward to employ a variable voltage source, but we need a considerable effort to design such a compact voltage source as to be added to every neuron. Instead, we may prepare several fixed voltages and multiplex them to the comparator according to neuronal activity. This is the motivation of this study. What we are interested in are (i) whether or not stepwise control of the threshold voltage is effective for the IP function in a spiking RNN (SRNN) for temporal data learning and (ii) if it is, how far we can go in reducing the number of the voltage lines.

When we introduce variable $V_{thr}$, we need to care about SP for hardware design. With regard to SP implementation, spike-timing dependent plasticity (STDP) (Legenstein, et al., 2005; Ning, et al., 2015; Srinivasan, et al., 2017) is the most popular synaptic update rule. STDP is a comprehensive synaptic update rule that obeys Hebb's law, but it is not hardware-friendly; it requires every synapse to have a mechanism to measure elapsed time from arrival of a spike. Alternatively, we employ spike-driven synaptic plasticity (SDSP) (Brader et al., 2007; Mitra et al., 2009; Ning et al., 2015; Frenkel et al., 2019; Gurunathan and Iyer, 2020; Payvand et al., 2022; Frenkel et al., 2023) which is much more convenient for hardware implementation. It is a rule where an incoming spike change the synaptic weight depending on whether $V_{mem}$ of the post-synaptic neuron is higher than a threshold $V_{Lthr}^{UP}$ or lower than another threshold $V_{Lthr}^{DOWN}$. The magnitude relationship $V_{Lthr}^{DOWN} \leq V_{Lthr}^{UP} < V_{thr}$ is essential for correct learning hence $V_{Lthr}^{DOWN}$ and $V_{Lthr}^{UP}$ should be defined according to $V_{thr}$.

In this work we study an SRNN with IP and SP where $V_{thr}$, $V_{Lthr}^{UP}$ and $V_{Lthr}^{Down}$ are discretized and synchronized. In order to make our model hardware-oriented, synaptic weights $W$ are also discretized so that we can assume conventional digital memory circuits for storing weights. We perform simulations of learning and anomaly detection tasks for publicly available electrocardiograms (ECGs) (Liu et al., 2013; Kiranyaz et al., 2016; Das et al., 2018; Amirshahi et al., 2019; Bauer et al., 2019; Wang et al., 2019) and show the effectiveness of our model. In particular, we discuss how much we can reduce the discretized levels of $V_{thr}$ and $W$, which is an essential aspect for hardware implementation.

## 2 Methods

### 2.1 LIF neuron model

The neuron model we employ in this work is the LIF model (Holt and Koch, 1997), which is one of the best-known spiking neuron models due to its computational effectiveness and mathematical simplicity. The membrane potential $V_{mem}^i$ of neuron $i$ is given as

$$C \frac{dV_{mem}^i}{dt} = I_{in} - \frac{V_{mem}^i}{R}$$

where $C$, $R$ and $I_{in}$ denote the membrane capacitance, resistance, and the sum of the input current flowing into the neuron, respectively. If $V_{mem}^i$ exceeds the firing threshold $V_{thr}^i$, neuron $i$ fires and transfers a spike signal to the next neurons connected via a synapse. Then, neuron $i$ resets $V_{mem}^i$ to $V_{reset}$ and enters a refractory state for time $t_{ref}$, during which $V_{mem}^i$ stays at $V_{reset}$ regardless of $I_{in}$. The LIF neuron is hardware-friendly because it can be implemented in analog circuits using industrially manufacturable complementary-metal-oxide-semiconductor (CMOS) devices (Indiveri et al., 2011), as illustrated in Figure. 1A.

### 2.2 Synapse and SDSP

A synapse receives spikes from neurons and external input nodes. When a spike comes, a synapse converts the spike into a synaptic current $I_{syn}$ proportional to $W$ defined as

$$\tau_{syn} \frac{dI_{syn}}{dt} = -I_{syn} + \alpha W \delta(t - t_{spike}),$$

where $\tau_{syn}$ and $t_{spike}$ are a time constant, and $\alpha$ is an appropriately defined constant. This synapse model is also compatible with the CMOS design.

As mentioned above, we employ SDSP as the synaptic update rule for SP. The synaptic weight $W(i,j)$ between pre-synaptic neuron $i$ and post-synaptic neuron $j$ increases or decreases if $V_{mem}^j$ is higher or lower than the learning threshold $V_{Lthr}^{UP}(j)$ or $V_{Lthr}^{DOWN}(j)$ when the pre-synaptic neuron $i$ fires, as follows:

$$W(i,j) = \begin{cases} W(i,j) + LR_{SDSP} & if \quad V_{mem}^j > V_{Lthr}^{UP}(j) \\ W(i,j) - LR_{SDSP} & if \quad V_{mem}^j < V_{Lthr}^{DOWN}(j) \end{cases} \quad \text{when a spike arrives,}$$

where $LR_{SDSP}$ is the learning step, which is set to a constant value, as illustrated in Figure. 1B.

In practice, the range of $W$ is finite, $0 \leq W \leq W_{max}$, hence $LR_{SDSP}$ defines the resolution of $W$. Higher resolution is favorable for better performance in general, but this leads to a larger circuit area for storing $W$ values. Emerging memory elements such as memristors and phase change memory devices may be employed to avoid this issue (Lazar et al., 2009; Li and Li, 2013), but practical use of these emerging technologies is still a big challenge. In this work, we assume conventional CMOS digital memory cells for storing $W$, raising our interest in how much we can reduce the resolution of $W$ for practical application task. In this view, we discuss the feasibility of binary $W$, which is ideal for hardware implementation, later in this work.

A circuit that determines whether $W(i,j)$ should be potentiated, depressed, or unchanged can be designed with two comparator circuits; the one compares $V_{mem}^j$ with $V_{Lthr}^{UP}(j)$ and the other with $V_{Lthr}^{DOWN}(j)$ (see supplementary materials). Note that it is sufficient for each neuron to have a determinator; it is not necessary for each synapse to have it.

## 2.3 Event-driven stepwise IP

The IP model we employ executes a stepwise change of the firing threshold voltage $V_{thr}^i$ of neuron $i$ in an event-driven manner as

$$V_{thr}^i = \begin{cases} V_{thr}^i + LR_{thr} & if \quad C_{fire}^i > (1+\sigma/2)C_{IP} \\ V_{thr}^i - LR_{thr} & if \quad C_{fire}^i < (1-\sigma/2)C_{IP} \end{cases} \quad \text{when neuron } i \text{ fires,}$$

where $LR_{thr}$ denotes the changing step of $V_{thr}^i$ in a single IP operation, $C_{fire}^i$ a parameter that measures of the activity of neuron $i$, $C_{IP}$ a constant corresponding to the target activity. $\sigma$ is a parameter that defines a healthy regime of $C_{fire}^i$, $(1-\sigma/2)C_{IP} < C_{fire}^i < (1+\sigma/2)C_{IP}$, where IP operation is not executed (see supplementary materials for details) (Payvand et al., 2022). $C_{fire}^i$ is often referred to as a calcium potential (Brader et al., 2007; Indiveri and Fusi, 2007; Qiao et al., 2015), defined as

$$\tau_{IP} \frac{dC_{fire}^i}{dt} = -C_{fire}^i + \sum_{Firings\ of\ Neuron\ i} \delta(t - t_{fire}^i),$$

where $\tau_{IP}$ is a constant and $t_{fire}^i$ represents all the firing times of neuron $i$ (note that all the firing times are summed up). The behavior of $C_{fire}^i$ is illustrated in Figure.1C, showing that it can be used as an indicator of the neuron activity if the threshold $C_{IP}$ is appropriately determined.

The firing threshold of a LIF neuron is given as a reference voltage applied to a comparator connected to the membrane capacitor. Stepwise change of $V_{thr}^i$ is advantageous for hardware implementation because we do not need to design a compact voltage source circuit that can tune the output continuously. Instead, we need to prepare several fixed voltage lines and select one of them using a multiplexer, which is not a difficult task.

**2.4 Synchronization of IP and SP thresholds**

If the SDSP thresholds $V_{Lthr}^{UP/DOWN}$ are fixed to be constants, the IP rule introduced above interfere with SP because it changes the magnitude relationship between $V_{Lthr}^{UP/DOWN}$ and $V_{thr}$. For example, let us assume that $V_{thr}$ is lowered by IP and comes below $V_{Lthr}^{DOWN}$. In this case, $W$ decreases every time a spike comes and finally reaches zero because $V_{mem}$ is always less than $V_{Lthr}^{DOWN}$ and never exceeds $V_{Lthr}^{UP}$. This would lead to incorrect learning of the input information.

To operate both IP and SP at the same time correctly, we synchronize the three thresholds of neuron $i$, that is, $V_{thr}^i$, $V_{Lthr}^{UP}(i)$, and $V_{Lthr}^{DOWN}(i)$ so that the magnitude relationship $V_{Lthr}^{DOWN} < V_{Lthr}^{UP} < V_{thr}$ should be kept during IP operations. Along with the firing threshold $V_{thr}^i$, the learning thresholds $V_{Lthr}^{UP}(i)$ and $V_{Lthr}^{DOWN}(i)$ are updated by IP as follows,

$$V_{Lthr}^{UP/DOWN}(i) = \begin{cases} V_{Lthr}^{UP/DOWN}(i) + LR_{Lthr}^{UP/DOWN} & if\ C_{fire}^i > (1+\sigma/2)C_{IP} \\ V_{Lthr}^{UP/DOWN}(i) - LR_{Lthr}^{UP/DOWN} & if\ C_{fire}^i > (1-\sigma/2)C_{IP} \end{cases} \text{when neuron } i \text{ fires,}$$

where $LR_{thr}^{UP/DOWN}$ are the change width of the learning thresholds.

**2.5 Network Model**

Figure 2A shows the architecture of the SRNN system we study in this work. It consists of an input layer, a middle layer, and an output layer. The middle layer (M-SRNN) is an RNN with random connections and synaptic weights, consisting of two neuron types which are excitatory and inhibitory neurons. The M-SRNN in this work consists of 80% excitatory and 20% inhibitory neurons. Input-layer neurons send Poisson spikes to the neurons of the M-SRNN at a frequency corresponding to the value of the input data. The input-layer neurons connect with excitatory neurons of M-SRNN with a probability of $P_{in}$, which is 0.1 in this work. Note that they have no connections to inhibitory neurons. The excitatory neurons connect with other excitatory neurons with probability $P_{EE}$ and with inhibitory neurons with probability $P_{EI}$. Inhibitory neurons connect with excitatory neurons with probability $P_{IE}$ and do not connect with inhibitory neurons. Output-layer neurons are connected from all excitatory neurons of M-SRNN. Not all M-SRNNs will give the desired result because of the random nature, so parameters related to the structure of M-SRNN must be set carefully to obtain the desired results (Payvand et al., 2022). With self-organization mechanism by IP and SP, the M-SRNN reconstruction is automatically performed using spike signals from input layer neurons.

The M-SRNN can be implemented as a crossbar architecture (Lazar et al., 2009) shown in Figure 2B. There, each row line is connected to a neuron of the M-SRNN, and each column line is connected to either an input-neuron emitting spikes in response to external inputs or a recurrent input from an M-SRNN neuron. A cross point is a synapse, where spikes from the column line are converted to synaptic current flowing into the row line. Some of the synapses are set inactive to realize the random connection of the RNN.

**3    Simulation and Results**

## 3.1 Simulation configuration and parameters

The effectiveness of our SRNN model with IP and SP explained above is evaluated using Brian simulator (Goodman and Brette, 2008) by ECG anomaly detection benchmark (PhysioNet 1999; Goldberger et al., 2000; Moody and Mark, 2001) with parameters listed in Table 1. Input-layer neurons convert the ECG data to Poisson spikes and send them to excitatory neurons in the M-SRNN. The simulation consists of three phases. Phase 1 is the unsupervised learning phase of the M-SRNN by using the training data of the ECGs. Thresholds ($V_{thr}$, and $V_{Lthr}^{UP/DOWN}$) of excitatory neurons and synaptic weights ($W$) between excitatory neurons in the M-SRNN are learned by IP and SP, respectively. Phase 2 is a readout learning phase. Synaptic weights between neurons in the M-SRNN and those in the output layer are calculated by linear regression in a supervised fashion. Phase 3 is the test phase. Using test ECG data, anomaly detection performance of the SRNN determined in Phase 1 and Phase 2 is evaluated.

In the simulation, the learning step $LR_{SDSP}$ and the firing threshold change width $LR_{thr}$ are selected from $S_{LR} = \{0.1, 0.2, 0.5, 1.0, 2.0\}$ and $P_{thr} = \{0.025\,V, 0.05\,V, 0.1\,V, 0.3\,V\}$, respectively. The ranges of $W$ and $V_{thr}$ are $0 \leq W \leq 2$ and $0.1\,V \leq V_{thr} \leq 0.4\,V$. With regard to the SP synchronization with IP, we set $V_{Lthr}^{UP}(i) = V_{Lthr}^{DOWN}(i) = V_{thr}^{i}/2$ throughout this work, hence $LR_{Lthr}^{UP} = LR_{Lthr}^{DOWN} = LR_{thr}/2$. All initial synaptic weights between excitatory neurons are set to 1.0, and the initial firing threshold is set to 0.2V for all neurons. All other synaptic weights are set randomly.

## 3.2 ECG anomaly detection

For ECG anomaly detection, we use the MIT-BIH arrhythmia database (PhysioNet 1999; Goldberger et al., 2000; Moody and Mark, 2001). Using the PhysioBank ATM provided by PhysioNet (PhysioNet 1999), we download and use MIT-BIT Long-Term ECG number 14046 for performance evaluation. Figure 3A shows normal waveform of the ECG used as training data. As test data, we use waveform data that partially include multiple abnormal waveforms, as shown in Figure 3B. To perform anomaly detection, the SRNN is used as an inference machine. Values of the data points of the ECG waveform are inputted to the SRNN one by one in the time order. At the $k$-th input, it predicts the next $(k + 1)$-st. The firing frequency $F_{out}(k)$ of the output-layer neuron at the $k$-th input is compared to the firing frequency of the input neuron at the $(k + 1)$-st input $F_{in}(k + 1)$. When the absolute difference $D(k + 1) = |F_{out}(k) - F_{in}(k + 1)|$ is greater than the abnormality judgment threshold $F_{thr}$, the $(k + 1)$-st input data is regarded to be abnormal.

Since the raw ECG data $E_{input}$ is given by time-series data of electrostatic potential in $mV$, the input-layer neurons convert the potential $E_{input}$ to the firing frequency $F_{in}$ as follows,

$$F_{in}(k) = F_{poisson} \times \frac{4 + 2 \times E_{input}(k)}{5}.$$

where $F_{poisson}$ is the maximum frequency. Since an input-layer neuron fires with Poisson probability $F_{in}(k)$, a single input is required to be kept for a certain duration ($T_{bin}$) to generate a desired Poisson spike train.

For correct abnormality detection, the abnormality judgment threshold $F_{thr}$ must be set within a range that is larger than the maximum difference $D_{no}^{max}$ for normal data input and smaller than the minimum difference $D_{ab}^{mim}$ for abnormal data input. The threshold width $W_{thr} = D_{ab}^{mim} - D_{no}^{max}$ represents flexibility. If this threshold width is large, the judgment capability is high because it allows a margin for judgment, and if it is small, the judgment capability is low because it increases the possibility of misjudgment.

### 3.3 Simulation Results

*Reduction of parameter resolutions toward hardware implementation*

Figure 4 shows anomaly detection results of the original M-SRNN, the M-SRNN reconstructed with SP and that with both SP and IP. Figures 4A, C and E show the difference $D(k)$ in the three cases respectively, when a normal ECG waveform in Figure 4G is used as a test data. Figures 4B, D, and F show the difference $D(k)$ when an abnormal ECG waveform in Figure 4H is used. It can be seen in Figure 4B that the initial M-SRNN cannot detect anomalous points because $D_{ab}^{mim}$ is almost the same as $D_{no}^{max}$, hence no window $W_{thr}$. On the other hand, the M-SRNN reconstructed by SDSP shows clear $W_{thr}$, as shown in Figure 4D. Furthermore, if we add the IP function, we obtain larger $D_{ab}^{min}$ while $D_{no}^{max}$ is the same, hence larger $W_{thr}$. This result clearly shows the effectiveness of the proposed SP and IP models in improving the anomaly detection performance. In particular, if the M-SRNN is reconstructed by using the synchronized SP and IP, a sufficient margin is obtained for anomaly detection without misdetection of normal data. Figure 5 shows a heat map of $W_{thr}$ at each $LR_{SDSP} \in S_{LR}$ and $LR_{thr} \in P_{thr}$ when the processing time $T_{bin}$ per one ECG data point for reconstruction is set to be $7\ ms$ (A), $150\ ms$ (B) and $600\ ms$ (C). These figures show that $W_{thr}$ becomes large as the operation time $T_{bin}$ increases, which is a reasonable result because the longer $T_{bin}$ becomes, the more information is learned from the data point, leading to higher accuracy of the abnormal detection. In fact, as can be seen in Figure 6, which shows $D(k)$ patterns for an abnormal waveform obtained with the M-SRNN reconfigured by $LR_{SDSP} = 0.1$ and $LR_{thr} = 0.3\ V$ for each $T_{bin}$, $D(k)$ becomes smoother and $D_{no}^{max}$ lower as $T_{bin}$ is set longer.

*Real-time operation for practical applications*

For practical application, it is desired that the abnormal data should be detected at the moment it occurs and thus real-time operation is highly expected. In this sense, $T_{bin}$ is desired to be as short as possible. In the case of the ECG anomaly detection, data is collected at 128 steps/sec. Therefore, the learning process and anomaly detection must be performed within $T_{bin} = 7\ ms$. However, as discussed above, such short $T_{bin}$ leads to small $W_{thr}$ because the learning duration for each data point is insufficient.

Now we assume that employing longer $T_{bin}$ is equivalent to increasing the number of IP and SP operations within short $T_{bin}$. To increase the number of IP and SP operations, we have to enhance the activities of neurons, hence two options. The first one is to enhance the parallelism of the inputs; we increase the number of neurons in the input layer $N_{input}$ so that a neuron in the M-SRNN being connected to the input layer receive more spike signals during short $T_{bin}$. The other is to enhance the seriality of the input neuron signals; we increase the rate of Poisson spikes $F_{Poisson}$ from the input layer. The effects of these two methods are verified by simulation.

Figure 7 shows the heatmaps of $W_{thr}$ for $T_{bin} = 7\ ms$ in the cases of $N_{input} = 10, 100$ and $200$. We observe that $W_{thr}$ increases with $N_{input}$ in general, indicating that our first idea is effective; real-time anomaly detection without false positive detection is possible by increasing $N_{input}$. Note that the binary $V_{thr}$ and $W$ i.e., $LR_{SDSP} = 2.0$ and $LR_{thr} = 0.3\ V$ result in sufficiently large $W_{thr}$ even with $T_{bin} = 7\ ms$ in the case of $N_{input} = 100$. Thus, a highly parallelized input layer has been shown to be effective for performance improvement with short $T_{bin}$. However, when $N_{input}$ is increased too much, the effect would be negative. As can be seen in Figure 7C, where $N_{input} = 200$, the M-SRNN does not work appropriately when $LR_{SDSP} = 2.0$ and $LR_{thr} \geq 0.2\ V$. Since the M-SRNN neurons that receive input spikes are always very close to the saturation in the case of large $N_{input}$, precise control of the parameters such as $V_{thr}$ and $W$ is required.

To examine the latter idea, we perform the anomaly detection tasks with $F_{Poisson}$ being varied. In the center of Figure 8, we plot obtained $W_{thr}$ as a function of $F_{Poisson}$. If increasing $F_{Poisson}$ does not play an effective role on performance improvement, $W_{thr}$ increases just linearly with $F_{Poisson}$, as indicated by a red dotted line. As a matter of the fact, however, we obtain $W_{thr}$ above the red line up to $F_{Poisson} = 1200\ Hz$, indicating that raising $F_{Poisson}$ improves the anomaly detection performance of an M-SRNN.

We observe in Figures 8 A-C that increasing $F_{Poisson}$ elevates the base line of $D(k)$ and magnify the peaks. This is reasonable because the more input spikes come, the more frequently the neurons in the M-SRNN fire, hence $D(k)$ being scaled with $F_{Poisson}$. At the same time, it smoothens variation of $D(k)$, indicating improved learning performance due to the increased IP and SP operations. This results in $W_{thr}$ being larger than the red dotted line. When $F_{Poisson}$ is increased further to 1500 Hz, the peaks corresponding to the abnormal data in the original waveform saturate, as can be seen in Figure 8D. This is because of the refractory time of neurons. Since a neuron cannot fire faster than its refractory time, it has an upper limit in its firing frequency. The saturation observed in Figure 8D is interpreted as a case where the firing frequency at the anomaly data points reaches its limit. As a result, $W_{thr}$ at $F_{poisson} = 1500\ Hz$ is suppressed and comes below the red dotted line. This discussion can be clearly seen in Figure 9, which shows the evolutions of $D_{no}^{max}$ and $D_{ab}^{min}$ with $F_{poisson}$ of the input neurons. We observe that $D_{no}^{max}$ increases linearly, while $D_{ab}^{min}$ increases only up to $F_{poisson} = 1200 Hz$. For $F_{poisson} \geq 1200\ Hz$, $D_{ab}^{min}$ reaches its limit and only $D_{no}^{max}$ increases, hence smaller $W_{thr}$. We note that the results shown in Figure 8 are obtained with $LR_{SDSP} = 2.0$ and $LR_{thr} = 0.3\ V$ i.e., binarized $V_{thr}$ and $W$.

It is noteworthy that we have found that binary $V_{thr}$ and $W$ may be employed if the input layer is optimized. This is highly advantageous for hardware implementation. For $V_{thr}$ (and also for $V_{Lthr}^{UP/Down}$), we may prepare the smallest 2-input multiplexers and only two voltage lines (see supplementary materials). What is more conspicuous is that $W$ can be reduced to binary. This means that for synapses we have no need of using an area-hungry multi-bit SRAM array or waiting for analog emerging memories, but we may employ just small 1-bit latches (see supplementary materials). Since the number of synapses scales with square of the number of neurons, this result has a large impact on the SRNN chip size.

Thus, optimization of the input gives a large impact on both performance and physical chip size of the SRNN. Whether we optimize $N_{iput}$ or $F_{Poisson}$ may be up to engineering convenience. It is possible to optimize both. As we have seen in Figures 7 and 8, the former has a better smoothing effect in the normal data area than the latter. Considering hardware implementation, on the other hand, the latter is more favorable because the former requires physical extension of the input layer system. For the latter, we only have to tune the conversion rate of raw input data to spike trains, which may be done externally. Therefore, the parameters in the input layer should be designed carefully taking those conditions discussed above into consideration.

## 4   Discussion

Lazer et al. proposed to introduce two plasticity mechanisms, SP and IP, to an RNN to reconstruct its network structure in the training phase (Lazar et al., 2009). While software implementation of SP and IP seems to be quite simple, we need some effort for hardware implementation.

   With regard to the IP operation, Lazer et al. adjusted the firing threshold of each neuron according to its firing rate at every time step. In hardware implementation, constantly controlling the thresholds of all of the *N* neurons is not realistic. Therefore, we proposed a mechanism that regulate the threshold of a neuron in an event-driven way; each neuron changes its firing threshold when it fires in accordance with its activity being higher or lower than the predetermined levels. This event-driven mechanism releases us from designing a circuit for precise control of the thresholds. As discussed by Lazar et al., we need to control the thresholds with an accuracy of $1/1000$ if it is done constantly, which requires quite large hardware resource that consumes power as well. Our event-driven method, on the other hand, has been shown to allow us stepwise control of the thresholds with only a few gradations, which is highly advantageous for hardware implementation.

   Another way to realize the IP mechanism is to regulate the current of a LIF neuron (Holt and Koch, 1997). The current value can be adjusted by changing the resistance values in the previous researches (Dalgaty et al., 2019, Zhang et al., 2022). This can be achieved by using variable resistors such as memristors (Dalgaty et al., 2019, Payvand et al., 2022) or by selecting several fixed resistors prepared in advance. For the former method, precise control of the resistance would be a central technical issue, but it is still a big challenge even today because the current memristor has large variation (Dalgaty et al., 2019). Payvand et al. discussed that variation and stochasticity of rewriting may lead to better performance, but further studies including practical hardware implementation and general verification are yet to be done. The latter requires a set of large resistors ($\sim 100$ $M\Omega$) for each neuron, which is not favorable for hardware implementation because resistors occupy quite large chip area. We believe that stepwise change of the firing threshold is the most favorable implementation of IP.

   For implementation of the SP mechanism, STDP (Legenstein, et al., 2005; Ning, et al., 2015; Srinivasan, et al., 2017) is widely known as a biologically plausible synaptic update rule, but it is not hardware friendly as discussed in the introduction. Hence recent neuromorphic chips tend to employ SDSP (Brader et al., 2007; Mitra et al., 2009; Ning et al., 2015; Frenkel et al., 2019; Gurunathan and Iyer, 2020; Payvand et al., 2022; Frenkel et al., 2023). However, SDSP cannot be implemented concurrently with threshold-controlled IP in its original form, because the latter may push down the upper limit of the membrane potential (i.e., the firing threshold) below the synaptic potentiation threshold. Our proposal that the synaptic update thresholds synchronize with the firing threshold realized the concurrent implementation of the two, and their interplay with each other led to successful learning and anomaly detection of ECG benchmark data (PhysioNet 1999; Goldberger et al., 2000; Moody and Mark, 2001) even with binary thresholds and weights if the parallelism and the seriality of the input are well optimized. This is highly advantageous for analog circuitry implementation from the viewpoints of circuit complexity and size.

| | Neurons | | Synapses | |
|---|---|---|---|---|
| | Excitatory | Inhibitory | $W$ | 1.0 |
| # of Neurons | 160 | 40 | SRNN | |
| $R\ (M\Omega)$ | 400 | 400 | $P_{EE}$ | 5% |
| $C\ (pF)$ | 10 | 10 | $P_{II}$ | 0% |
| $\tau_{Ca}\ (ms)$ | 100 | 100 | $P_{EI}$ | 2% |
| $V_{th}\ (V)$ | 0.2 | 0,2 | $P_{IE}$ | 10% |

Table1. Initial values in SRNN simulations.

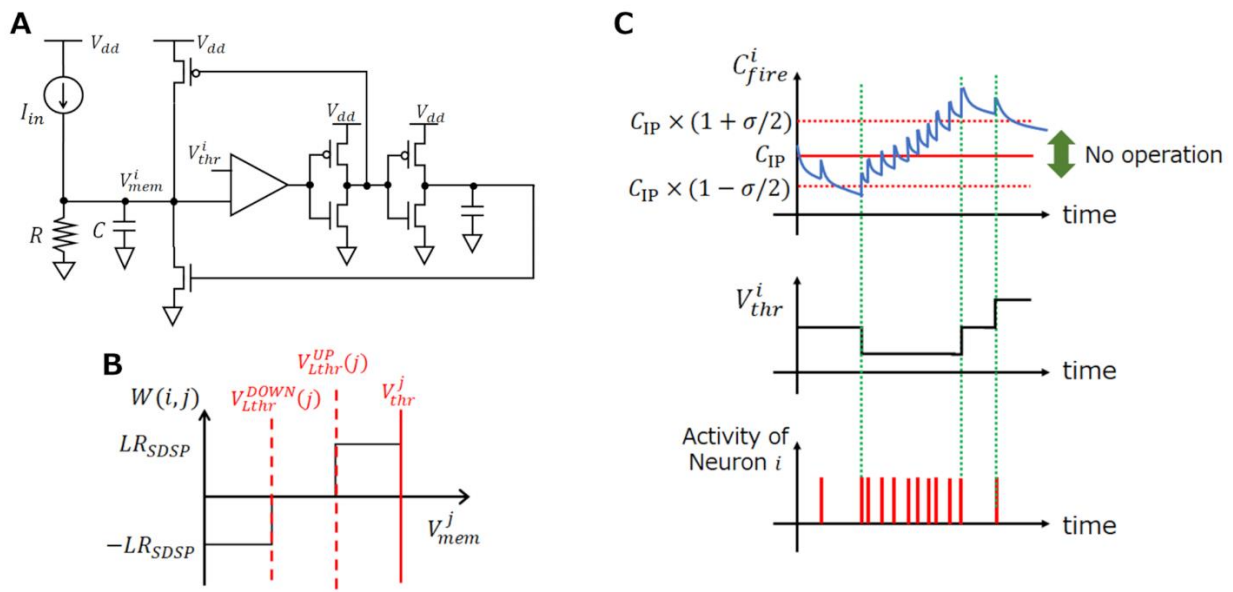

Figure 1. Model and behavior of each component of SRNN. **(A)** LIF neuron circuit diagram. **(B)** Schematic diagram of synaptic weight variation. **(C)** Behavior of $C_{fire}^i$ and $V_{thr}^i$ depending on $C_{IP}$.

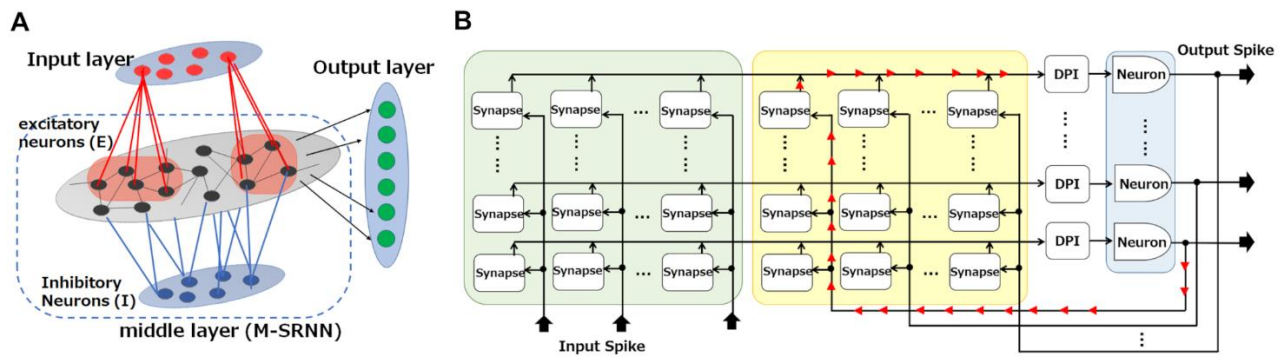

Figure 2. Hardware implementation for an SRNN. **(A)** SRNN consists of input, middle (M-SRNN), and output layers. The M-SRNN consists of excitatory (*E*, black) and inhibitory (*I*, blue) sub-population layers. **(B)** Hardware implementation for M-SRNN.

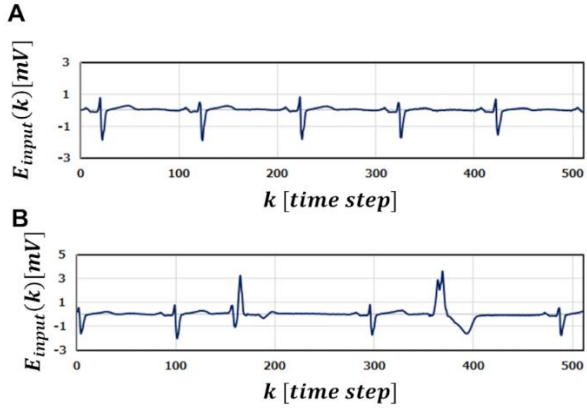

Figure 3. ECG benchmark waveform (number 14046) used in the simulation. **(A)** Normal ECG waveform. **(B)** Abnormal ECG waveform.

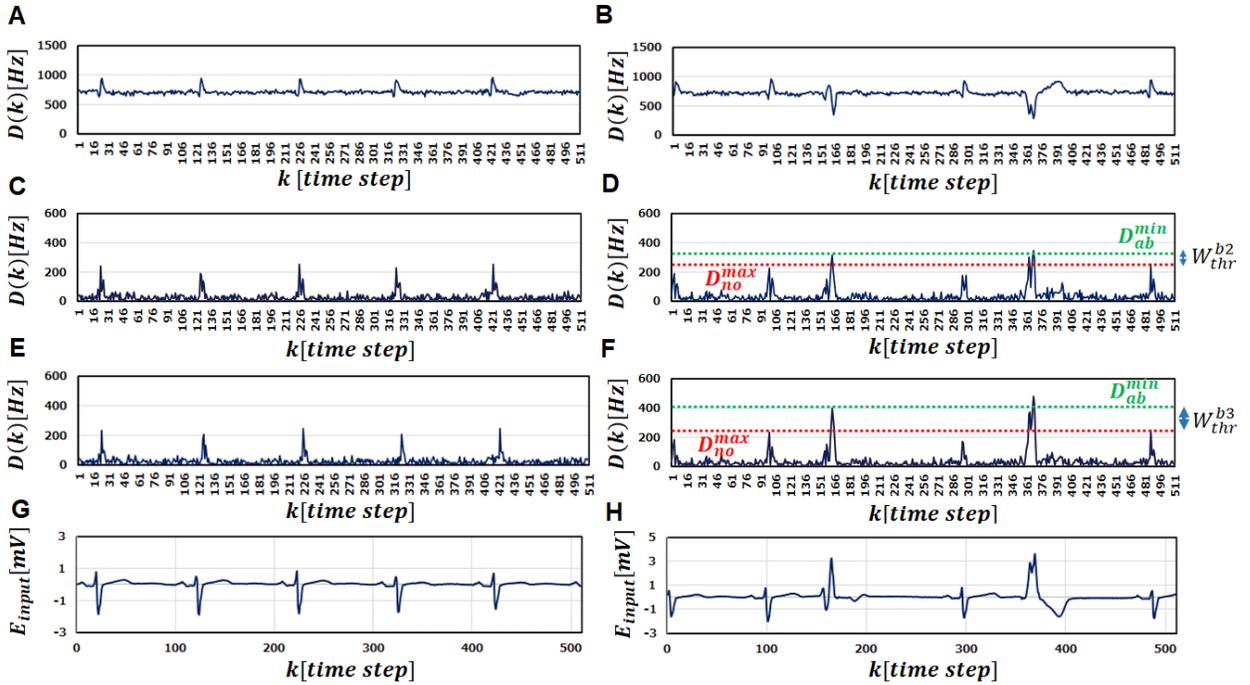

Figure 4. $D(k)$ with $T_{bin} = 150\ ms$, $LR_{SDSP} = 2.0$, $LR_{thr} = 0.025\ V$, and $\sigma = 0.3$. **(A)(B)** Normal and abnormal cases by initial M-SRNN, **(C)(D)** by M-SRNN reconstructed by only SDSP learning, **(E)(F)** by M-SRNN reconstructed by SDSP and synchronized IP learning, and **(G)(H)** original data waveform, respectively.

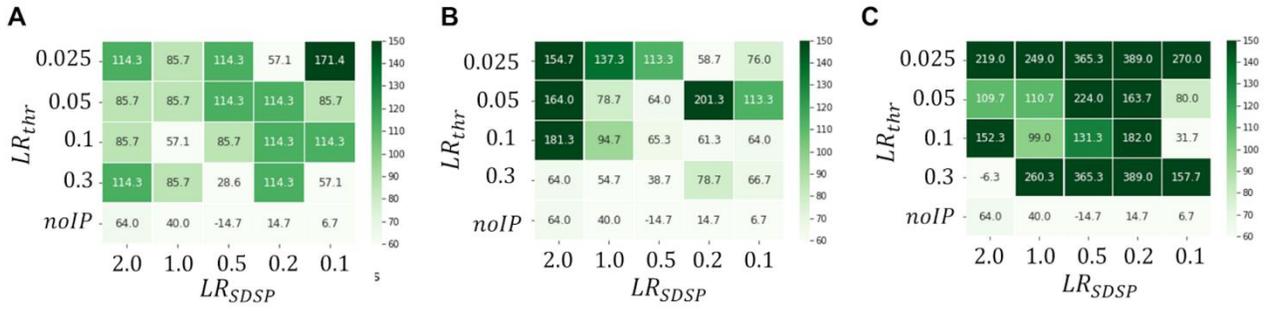

Figure 5. Heatmaps of $W_{thr}$. $T_{bin} =$ (**A**) $7\ ms$, (**B**) $150\ ms$, and (**C**) $600\ ms$, respectively.

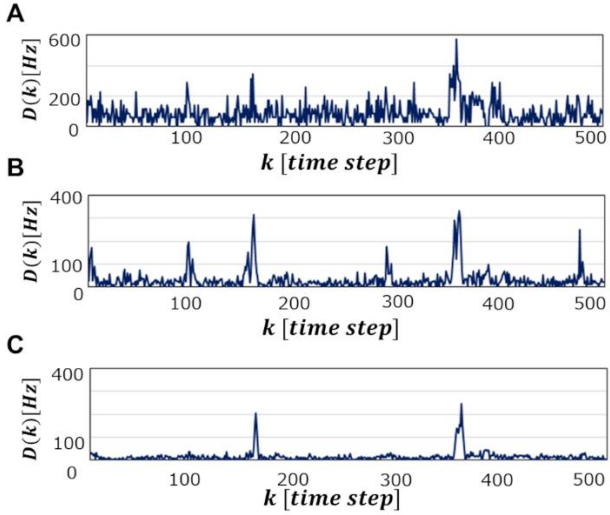

Figure 6. $D(k)$ when abnormal ECG waveform is detected in SRNN reconstructed with $LR_{SDSP} = 0.1$ and $LR_{thr} = 0.3V$. $T_{bin} =$ (**A**) $7\ ms$, (**B**) $150\ ms$, and (**C**) $600\ ms$, respectively.

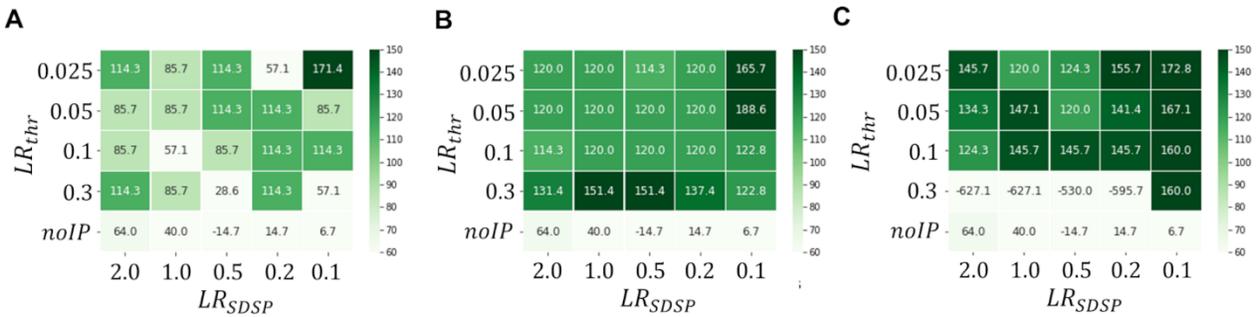

Figure 7. Heatmaps of $W_{thr}$ in case of $N_{input} =$ (**A**) $10$, (**B**) $100$, and (**C**) $200$ ($T_{bin} = 7ms$), respectively.

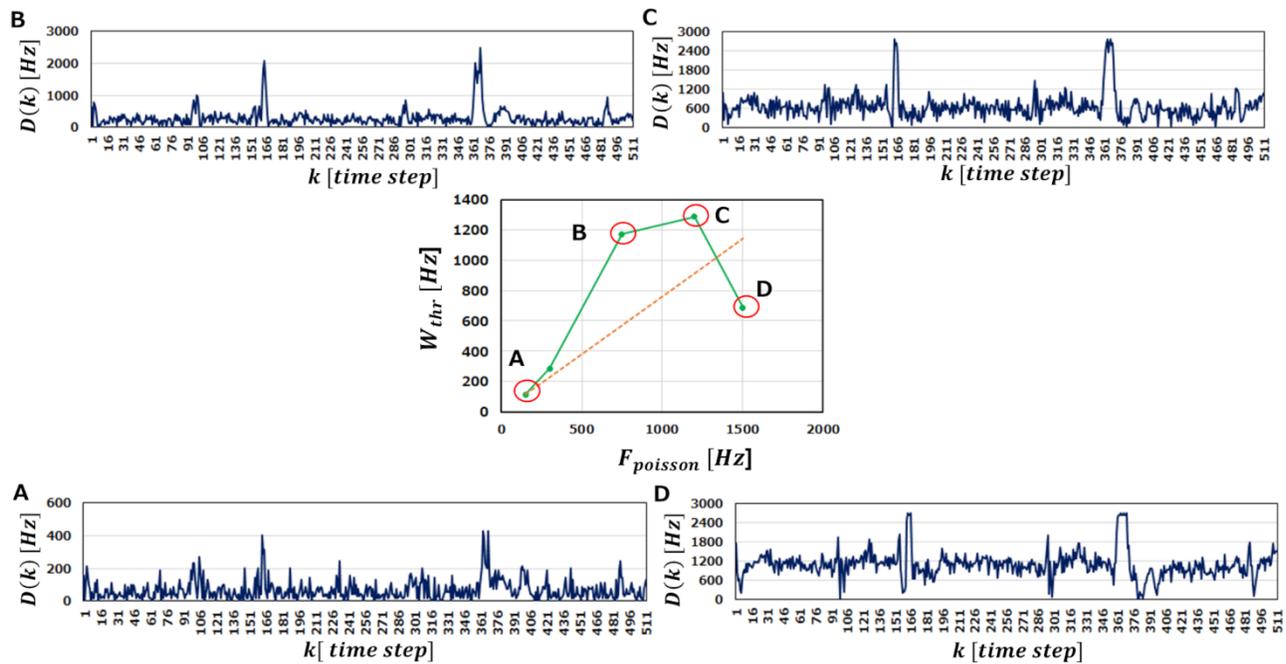

Figure 8. $W_{thr}$ and $D(k)$ in the case of $LR_{SDSP} = 2.0$ and $LR_{thr} = 0.3V$. The center graph shows the $W_{thr}$ against $F_{poisson}$. The outer diagrams represent $D(k)$ corresponding to **(A)-(D)** points in the center diagram. $F_{poisson} =$ **(A)** 150, **(B)** 750, **(C)** 1200, and **(D)** 1500$Hz$, respectively.

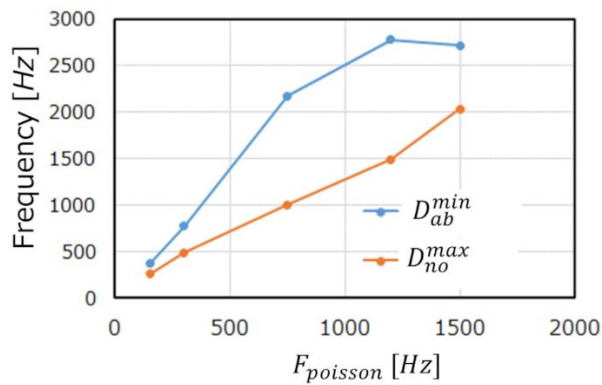

Figure 9. $D_{ab}^{min}$ and $D_{no}^{max}$ with $LR_{SDSP} = 2.0$ and $LR_{thr} = 0.3V$ against $F_{poisson}$.